\documentclass[12pt]{article}
\usepackage{graphicx}
\usepackage{times}
\newcommand{\ignore}[1]{}

\usepackage{amsmath,amssymb} 
\usepackage{color}
\usepackage{fullpage}

\usepackage{booktabs}
\usepackage{multirow}
\usepackage{siunitx,booktabs,caption}
\usepackage[pagebackref=true,breaklinks=true,letterpaper=true,colorlinks,bookmarks=false,citecolor=blue,linkcolor=blue,urlcolor=blue]{hyperref}
\usepackage{subfigure}
\usepackage{cite}
\usepackage{hyperref}

\usepackage[accsupp]{axessibility}  

\newcommand{\figref}[1]{Fig.~\ref{#1}}%
\newcommand{\tabref}[1]{Table~\ref{#1}}%
\newcommand{\secref}[1]{Sec.~\ref{#1}}
\renewcommand{\eqref}[1]{Equation~ (\ref{#1})}

\makeatletter
\def\thanks#1{\protected@xdef\@thanks{\@thanks
        \protect\footnotetext{#1}}}
\makeatother

\begin{document}

\title{FlowNAS: Neural Architecture Search for Optical Flow Estimation} 

\author{
	Zhiwei Lin$^{1}$\textsuperscript{*\thanks{\textsuperscript{*}Equal contribution.  \textsuperscript{\ddag}Contact person.}} \quad
	Tingting Liang$^{1}$\textsuperscript{*} \quad
	Taihong Xiao$^{2}$\textsuperscript{*} \\
	Yongtao Wang$^{1}$\textsuperscript{\ddag} \quad
	Zhi Tang$^{1}$ \quad
	Ming-Hsuan Yang$^{2}$
	\\
	\vspace{-1em}
	\and
	$^{1}$Wangxuan Institute of Computer Technology, Peking University\\
	$^{2}$University of California, Merced\\
	\vspace{-1em}
	{\tt\small \{zwlin,tingtingliang,wyt\}@pku.edu.cn\quad \{txiao3,mhyang\}@ucmerced.edu}\\
}
\date{}

\maketitle

\begin{abstract}
Existing optical flow estimators usually employ the network architectures typically designed for image classification as the encoder to extract per-pixel features. 
However, due to the natural difference between the tasks, the architectures designed for image classification may be sub-optimal for flow estimation. 
%
To address this issue, we propose a neural architecture search method named FlowNAS to automatically find the better encoder architecture for flow estimation task. 
We first design a suitable search space including various convolutional operators and construct a weight-sharing super-network for efficiently evaluating the candidate architectures. 
Then, for better training the super-network, we propose Feature Alignment Distillation, which utilizes a well-trained flow estimator to guide the training of super-network. 
Finally, a resource-constrained evolutionary algorithm is exploited to find an optimal architecture (i.e., sub-network). 
 Experimental results show that the discovered architecture with the weights inherited from the super-network achieves 4.67\% F1-all error on KITTI, an 8.4\% reduction of RAFT baseline, surpassing state-of-the-art handcrafted models GMA and AGFlow, while reducing the model complexity and latency. 
The source code and trained models will be released in \href{https://github.com/VDIGPKU/FlowNAS}{https://github.com/VDIGPKU/FlowNAS}.

\end{abstract}

\section{Introduction}
Optical flow estimation aims to measure per-pixel 2D motion between consecutive video frames, which is widely used in various tasks, e.g., action recognition\cite{opticalforaction}, object tracking~\cite{trackwithopticalflow,bxz155tracking}, and video understanding \cite{FortunBK15}. 
One key component for accurate optical flow estimation lies in constructing a discriminative cost volume from an effective feature extractor.

Recently, deep neural networks have been applied to  optical flow estimation by extracting more effective features~\cite{dosovitskiy2015flownet,pwcnet,TeedD20raft}. 
With the pyramid coarse-to-fine decoder architecture design, the flow estimation can be further refined. 
While most flow estimation methods mainly focus on constructing a better cost volume~\cite{vcn,lcv,Jiang2021DCVNetDC} or designing a more delicate decoder~\cite{jiang2021gma}, we show that the encoder architecture is also essential for two reasons. 
First, the cost volume and flow decoder all rely on the feature representation extracted by the encoder. 
A better encoder can provide a better feature representation for cost volume construction or flow regression. 
Second, the encoder occupies a considerable part of the flow networks regarding the number of parameters. 
For example, the proportion of the encoder parameters in the RAFT~\cite{TeedD20raft} model is 58\%.
Consequently, the encoder dominates the training process.
However, the recent state-of-the-art methods still adopt the encoders designed for image classification 
to extract the feature of the input images. 
Due to the natural difference of image classification and flow estimation tasks, the encoder directly adopted from classification models is not optimal. 
The feature representation given by the sub-optimal encoder architecture may limit the performance upper bound of flow estimation methods based on precise cost volumes. 
Thus, designing a good encoder architecture remains an open question in flow estimation.

\begin{figure}[!t]
	\centering
	\caption{Model parameters, GFLOPs, latency vs. KITTI F1-all error. FlowNAS achieves better accuracy-efficiency trade-offs than handcrafted approaches.
	}
	\includegraphics[width=1.0\linewidth]{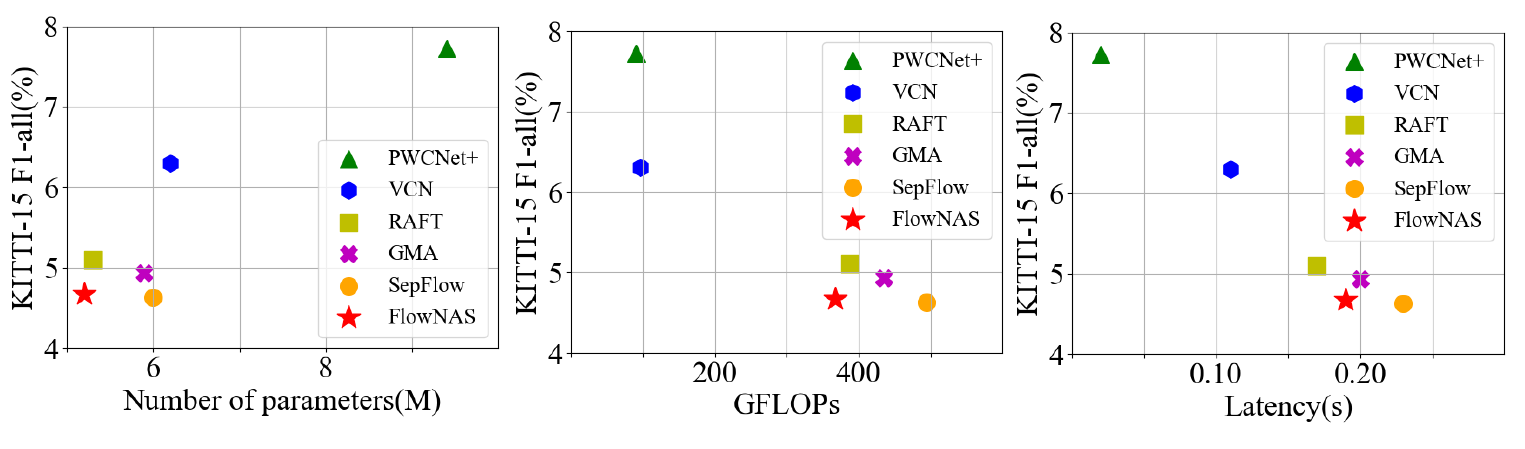}
	\label{fig:time}
\end{figure}

On a separate line of research, to reduce human efforts in designing neural networks, Neural Architecture Search (NAS) has been successfully applied to various high-level vision tasks \cite{BenderKZVL18,TanPL20,LiuCSAHY019}.
Nevertheless, much less attention has been paid to using NAS for low-level vision tasks. 
This can be attributed to several factors. 
In general, NAS requires selecting network components (e.g., the kernel size of convolution in a particular layer) through a large number of possible architectures. 
This entails heavy computational load (early NAS algorithms \cite{iclr/ZophL17} require thousands of GPU hours to find an architecture on the CIFAR dataset \cite{cifar}). 
In addition to the heavy computational requirements, existing NAS methods are designed for a specific task with different human priors.
%
To the best of our knowledge, NAS has not been employed to determine optimal encoders for flow estimation.  
%

In this paper, we present a neural network architecture search method, FlowNAS, specifically for better encoder design in optical flow estimation. 
We leverage the human knowledge in flow estimation as priors towards architecture search and design.
To handle the issue of computational overhead, we first determine a suitable search space by exploring the effectiveness of various convolutional operators and then construct a super-network that comprises all weight-sharing sub-network architectures.
We then propose the vanilla FlowNAS following \cite{abs-1904-00420} within two steps: (1) constraint-free super-network pre-training, (2) resource-constrained sub-network search. 
However, due to the large number of sub-networks sharing weights and thus interfering with each other, we find that the weights inherited directly from super-network are often sub-optimal. 
Hence, retraining the discovered architecture from scratch is usually required, introducing additional computational overhead. 

In contrast to existing NAS algorithms that use sophisticated pruning or sampling strategies \cite{CaiGWZH20,YuJLBKTHSPL20,Wang0GC21} to prevent interference between sub-networks, we propose Feature Alignment Distillation to fully utilize human priors in flow estimation and achieve much better performance. 
In particular, we take pre-trained weights of handcrafted optical flow estimator, e.g., RAFT \cite{TeedD20raft}, as our teacher model and super-network as our student model. 
In each update step, we sample one sub-network from super-network, and the extracted feature map pyramid is trained under the guidance of the teacher model. 
Specifically, Channel-wise Alignment is applied to the feature map of both teacher and student to make the distillation process independent of the number of channels. 
%
It is worth to mentioned that using existing open-source weights as the teacher guidance will not limit the performance of FlowNAS. Instead, it leads to the fast convergence and performance improvement of the super-network.

Therefore, we can simultaneously obtain outstanding model architectures and their corresponding weights by an evolutionary algorithm. 
Experimental results demonstrate that our searched model outperforms handcrafted models (\figref{fig:time}). 

\begin{figure}[!t]
	\centering
	\caption{Illustration of FlowNAS pipeline. It has two stages: (1) super-network training. (2) optimal architecture search. The searched architecture and its weights inherited from super-network can achieve better results than its stand-alone training.}
	\includegraphics[width=1.0\linewidth]{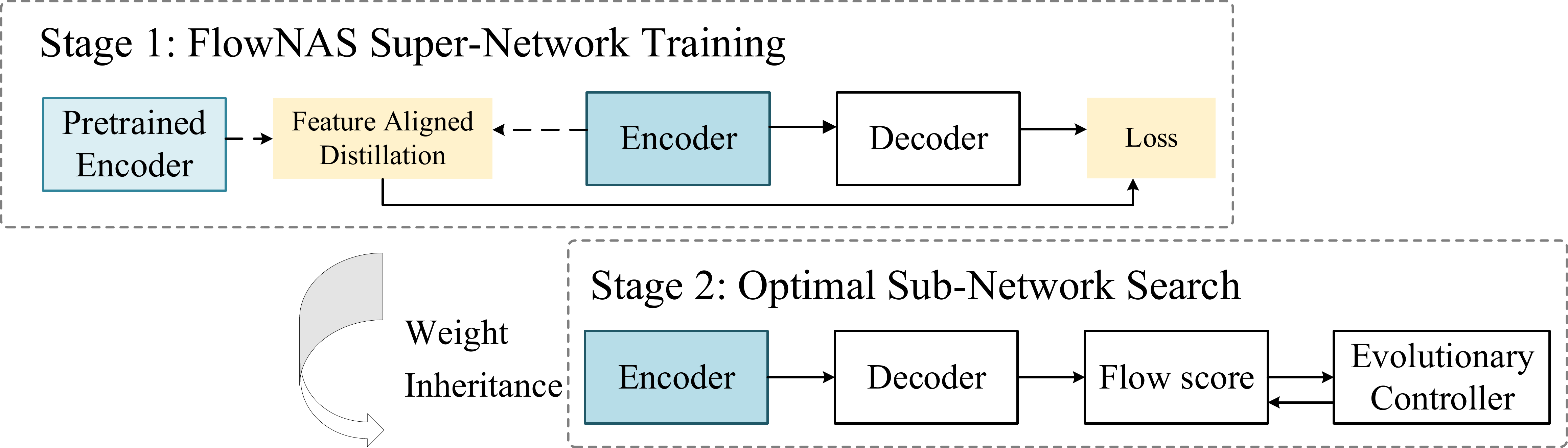}
	\label{fig:overall_pipe}
\end{figure}

The main contributions of this work can be summarized as:
\begin{itemize}
	\item We propose a neural architecture search framework, FlowNAS, specifically for optical flow estimation. 
	To the best of our knowledge, this is the first work for this challenging task.
	\item We analyze the importance of the encoder part of flow estimators and present an efficient search space.
	We propose Feature Alignment Distillation to achieve strong representation capability of super-network (\figref{fig:overall_pipe}) by effectively leveraging prior information.
	\item FlowNAS achieves the state-of-the-art accuracy-efficiency trade-offs on KITTI. 
	For example, FlowNAS-RAFT achieves an F1-all error of 4.67\% on KITT, an 8.4\% error rate reduction of RAFT \cite{TeedD20raft}, surpassing cutting edge handcrafted models including GMA \cite{jiang2021gma} and AGFlow \cite{agflow} while reducing the model complexity and latency.
\end{itemize}

\section{Related Work}

\subsection{Optical Flow Estimation}

Existing optical flow estimation methods have been inspired by the success of neural networks on per-pixel predictions. The first end-to-end deep neural network design for optical flow estimation is FlowNet~\cite{flownet}, where an encoder-decoder architecture is used. FlowNet2.0~\cite{flownet2} further extends FlowNet by stacking multiple basic FlowNet modules for iterative refinement. Motivated by the idea of a coarse-to-fine refinement paradigm, SpyNet~\cite{spynet} employs a spatial pyramid network that warps images at different scales to deal with large motions. PWC-Net~\cite{pwcnet}  extracts the feature through pyramid processing and builds a cost volume at each level, where the estimated flow gets iteratively refined. Aside from the pyramid architecture design, the cost volume has gotten more attention in recent works. VCN~\cite{vcn} improves the cost volume processing by decoupling the 4D convolution into a 2D spatial filter and a 2D winner-take-all filter. LCV~\cite{lcv} enhances the performance and robustness of flow estimation methods by introducing a learnable cost volume using Cayley representations. In \cite{9440793}, the authors perform grid search for locating the highest response in the spatiotemporal frequency domain to investigate how deep neural networks estimate optical flow. DCVNet~\cite{Jiang2021DCVNetDC} proposes dilated cost volumes to capture small and large displacements.
Moreover, another line of work on flow refinement is designing better decoder architecture. IRR~\cite{irr} first introduces an iterative residual refinement scheme. RAFT~\cite{TeedD20raft} further develop a lightweight recurrent decoder by sharing weights across the iterative refinement process. GMA~\cite{jiang2021gma} introduces a global motion aggregation to capture the long-range self-similarities in the reference frames. 

In contrast to the works above, where the model architecture is hand-crafted designed based on some principles, we aim to obtain better feature representations by searching for a better network architecture specifically for optical flow estimation.

\subsection{Neural Architecture Search}

\textbf{NAS for high-level vision tasks.}
Neural Architecture Search (NAS) is designed to replace the efforts of human experts in network architecture design by machines. 
It has achieved significant successes in numerous high-level vision tasks such as classification \cite{BenderKZVL18,BrockLRW18,abs-1904-00420,abs-1907-01845,LiuSY19,abs-1911-12126,CaiZH19}, object detection \cite{TanPL20,LiangWTHL21}, and semantic segmentation \cite{LiuCSAHY019,ZhangQLYLM19}. 
Early NAS methods \cite{LiuZNSHLFYHM18,LiuSVFK18,RealAHL19} train thousands of candidate architectures from scratch (on a smaller proxy task) and use their validation performance as feedback to an algorithm that learns to focus on the most promising regions in the search space. 
Recent works attempt to amortize the cost by training a single super-network, where the sub-networks can be efficiently ranked by using shared weights to estimate their relative accuracy \cite{BenderKZVL18,BrockLRW18,abs-1904-00420,abs-1907-01845}.  
To speed up the searching process, gradient-based approaches \cite{LiuSY19,abs-1911-12126} are proposed for continuous relaxation of the search space, which enables differentiable optimization in architecture search.  

All the above approaches require retraining: first training super-network to determine the optimal architecture (sub-network) configuration, and then retraining this architecture from scratch to obtain the final result, introducing computational overhead.
To alleviate these issues, several different pruning and training techniques~\cite{CaiGWZH20,YuJLBKTHSPL20,Wang0GC21} have been developed to improve the performance of super-network, where each sub-network performs on par with its stand-alone performance. 
For example, OFA \cite{CaiGWZH20} pre-trains a single whole network and then progressively distills it to obtain a smaller network. BigNAS \cite{YuJLBKTHSPL20} utilizes sandwich rule and inplace distillation to handle a wider set of models. 
AttentiveNAS \cite{Wang0GC21} uses a pareto front sampling strategy to better optimize super-network.

\vspace{1mm}
\noindent \textbf{NAS for low-level vision tasks.}
Advances of NAS have led to numerous applications in low-level vision tasks \cite{ZhangLCS20,LiTC20,ChengZHDCLDG20,Liu0Z0L21}.
Currently, gradient-based differentiable architecture search (DARTS)-type methods \cite{LiuSY19} are often used.
LEAStereo \cite{ChengZHDCLDG20} presents an end-to-end NAS framework for deep stereo matching by incorporating task-specific human knowledge into the architecture framework. 
In this method, a task-specific architecture search space is designed, and the differentiable optimization is used for architecture search. 
HiNAS \cite{ZhangLCS20} uses primitive search space (e.g., $3\times 3$ separable convolution) to address synthetic Gaussian noise removal for image restoration. %
RUAS \cite{Liu0Z0L21} designs an unrolling-type architecture search to handle low-light image enhancement. 
The searched architectures in the above methods perform well when retrained on specific datasets.
We note that existing NAS methods for low-level vision mainly require two-stage training: searching architecture and retraining. 
The full potential of NAS can be further exploited. 
%
To the best of our knowledge, NAS has not been employed to determine model architectures for flow estimation.

\section{NAS for Optical Flow Estimation}
In this section, we present our NAS optical flow estimation network. 
Our search method benefits from human knowledge in flow estimation.
We construct a compact and efficient search space by leveraging task-specific human knowledge. 
By utilizing the pre-trained weights of the handcrafted architecture, our trained super-network can achieve strong representation without requiring additional retraining.
\figref{fig:overall_pipe} shows the overall pipeline of FlowNAS.

\subsection{Problem Formalization}
Assuming the weights of super-network as $W$ and the architectural configurations as $\{\alpha_i\}$, we  formulate the problem as:
\begin{equation}
\begin{split}
    & \alpha = \mathop{\arg\min}\limits_{\alpha_i\in \mathcal{A}} Error_{val} (C (W, {\alpha_i})), \\
    & s.t. ~ Param (\alpha_i) < P, ~\forall i.\\
\end{split}
\end{equation}
where $C (W, {\alpha_i})$ denotes a selection scheme that selects part from super-network $W$ to form a sub-network with architectural configuration $\{\alpha_i\}$, and $P$ is the parameter upper bound.
The overall selecting scheme aims to find the sub-network with the lowest error that satisfies the resource constraint. 
The overall training objective is to optimize $W$ to make each supported sub-network maintain the same error rate level as its stand-alone performance.
\subsection{Search Space}
\label{search space}

\noindent\textbf{Encoder Design.}
We note that the deep models for optical flow estimation usually opt for an encoder-decoder structure \cite{TeedD20raft,jiang2021gma,Zhang2021SepFlow}. 
While most flow estimation methods focus on constructing a better cost volume or designing a more delicate decoder, we revalue the importance of encoder design for extracting basic features. 
Thus, we intend to extend NAS to search for better representative encoders in optical flow estimation, while integrating and optimizing different decoders in one super-network is not the focus of this paper. 
Motivated by the success of RAFT \cite{TeedD20raft} and CNN models \cite{HeZRS16,XieGDTH17}, we divide a CNN model into a sequence of units with gradually reduced feature map size and increased channel numbers. 
Each unit consists of a sequence of convolution layers. Following the common practice of NAS \cite{LiuSY19,abs-1904-00420}, we treat each unit as a searching cell, and the entire search space is a regular stack of these searching cells.

\vspace{1mm}
\noindent\textbf{Cell level search space.}
We allow each searching cell to use arbitrary numbers of convolution layers (denoted as dynamic depth) and each layer to use arbitrary numbers of channels (denoted as dynamic width) and arbitrary kernel sizes (denoted as dynamic kernel size). 
For example, the depth of each cell is chosen from $\{1, 2,3,4\}$. While for each layer, the width ranges from 56 to 136 with a stride 8, the kernel size is chosen from $\{3,5\}$, and the expansion ratio is chosen from $\{1, 2,4,5,6\}$.
With 6 cells, we have roughly $ ( (10\times 2 \times 5)^1 + (10\times 2 \times 5)^2 + (10\times 2 \times 5)^3 + (10\times 2 \times 5)^4)^6 	\approx 10^{48}$ different neural network architectures. 
Since all these sub-networks share the same weights from $W$, we only require 8.1M parameters to store all of them. 
Without sharing, the total model size will be computationally prohibitive.

We summarize the essential convolution operations for dense image prediction into three types: standard convolution, separable convolution \cite{Chollet17}, and shuffle convolution \cite{ZhangZLS18}. 
A straightforward approach is to combine them all in the search space, which will be expanded by over 1000 times, further increasing the difficulty of optimization of super-network. 
%
%
Instead, we conduct a simple comparison of these convolutions and choose the most suitable one, separable convolution, for optical flow estimation.
More details can be found in \secref{sec:ablation}.

\subsection{Vanilla FlowNAS}
Training super-network can be posed as a multi-objective problem, where each objective corresponds to one sub-network. 
A naive training approach is to directly optimize super-network by enumerating all sub-networks in each update step. 
However, it is computationally prohibitive for a vast search space like ours. 
Another naive training approach is to sample one sub-network in each update step, which prevents the issue of prohibitive cost. 
Following SPOS \cite{abs-1904-00420}, we propose vanilla FlowNAS, which has two steps: (1) \textit{constraint-free pretraining} - for each update step, we randomly sample one sub-network with architecture configuration $\{\alpha_i\}$ and update its weights in super-network $W$ by back propagation.
(2) \textit{resource-constrained search} - identifying the best-performed sub-networks under given resource constraint by an evolutionary algorithm. 
However, with such a large number of sub-networks sharing weights and thus interfering with each other, we find that weights directly inherited from super-network are often sub-optimal \cite{abs-1904-00420,abs-1907-01845}. 
Hence, retraining the discovered architecture from scratch is usually required, introducing additional computational overhead. 
In the following, we introduce a solution to address this challenge, i.e., \textit{Feature Alignment Distillation}.

\subsection{Feature Alignment Distillation}
A super-network comprises numerous sub-networks of different sizes. 
Previous works adopt sophisticated training strategies \cite{CaiGWZH20,YuJLBKTHSPL20,Wang0GC21} to prevent interference among sub-networks. 
However, applying them directly to flow estimation has two limitations: (1) it requires 2 to 3 times more training time to pre-train or sample more than one sub-network in each update step, (2) such training strategies are designed for image classification and may not be optimal for dense image prediction such as flow estimation.
More analysis can be found in \secref{sec:ablation}.

\begin{figure}[!t]
	\centering
	\caption{
	Overview of FlowNAS super-network training with Feature Alignment Distillation (FAD). It uses pre-trained weights of handcrafted flow estimator (below) as a teacher to guide super-network (above) training.  }
	\includegraphics[width=1.0\linewidth]{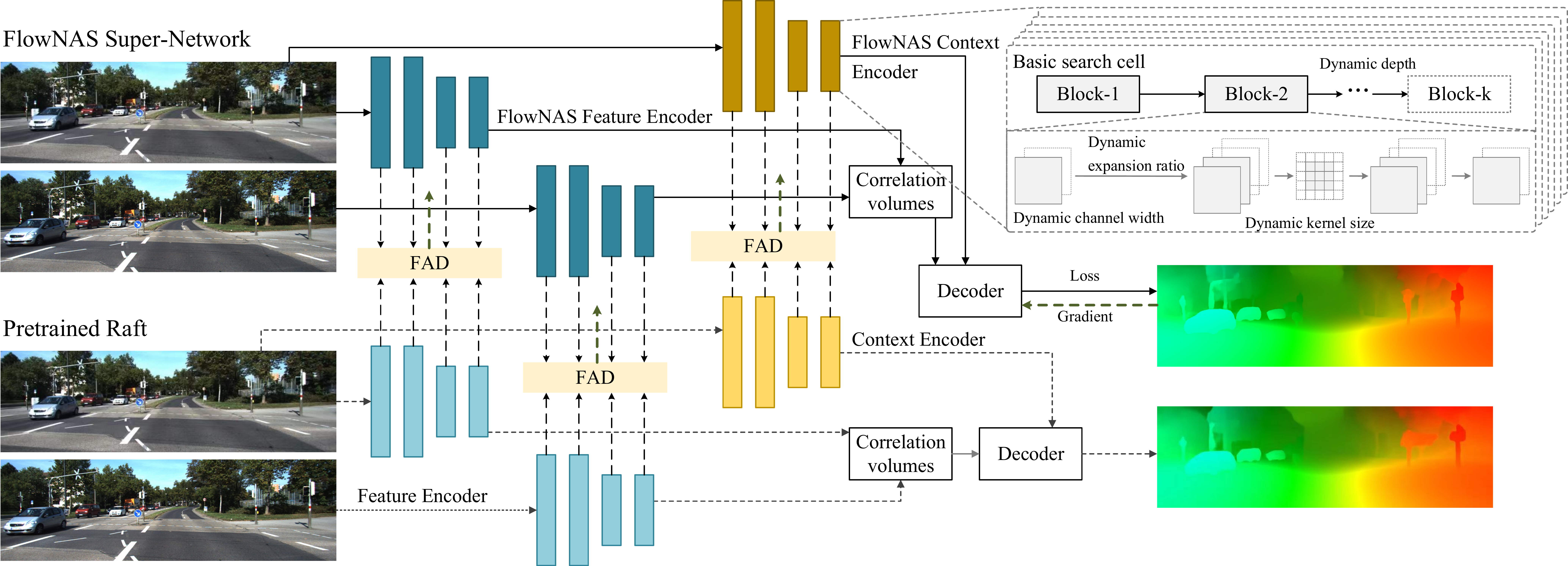}
	\label{fig:train_pipe}
\end{figure}

As our goal is to enhance the representation ability of super-network and prevent interference among sub-networks, it is natural to think of exploiting human knowledge to guide the training process. 
Instead of designing complex transfer rules, we propose using pre-trained weights of a handcrafted flow estimator as a teacher to guide our super-network training, thereby making the best use of human priors.
To this end, we propose Feature Alignment Distillation (FAD).
FAD can guide the feature outputs of all sub-networks sampled from super-network to converge into the same feature space, which reduces the training instability of the decoder and thus facilitates the training of the whole super-network.

As shown in \figref{fig:train_pipe}, FlowNAS takes a handcrafted estimator with the pre-trained weights, e.g., RAFT \cite{TeedD20raft}, as teacher model, and our super-network as student model. 
The feature pyramid extracted by teacher model and that of sub-network sampled from super-net are denoted as $\{f^T_i\in \mathcal{R}^{c^T_i\times h\times w}|i=1,2,...,n\}$ and $\{f^S_i\in \mathcal{R}^{c^S_i\times h\times w}|i=1,2,...,n\}$. 
Since the sub-networks have dynamic width in our search space, the channel number of $f^S_i$ is variable and not aligned with $f^T_i$. 
Thus, we apply a channel-wise alignment operation $g (\cdot)$ to $f^T_i$ and $f^S_i$ to align the number of the channel of two features. 
Then, we perform $L_2$ distance to estimate the difference between $g (f^T_i)$ and $g (f^S_i)$. \par

\vspace{1mm}
\noindent\textbf{Distillation Supervision:} The distillation loss computes the $L_2$ distance of $g (f^T_i)$ and $g (f^S_i)$ with exponentially increasing weight $\gamma^{N-i}$: 
\begin{equation}
    \mathcal{L_D}=\sum_{i=1}^N \gamma^{N-i}\cdot L_2 (g (f^T_i), g (f^S_i)).
\end{equation}
The final training loss for FlowNAS is: 
\begin{equation}
    \mathcal{L} = \mathcal{L}_{flow} + \lambda\mathcal{L_D},
\end{equation}
where $\mathcal{L}_{flow}$ is the original loss function for the flow method, and $\lambda$ is the weight to balance two losses. $\lambda$ is set to $1$ in our experiments.\par

\vspace{1mm}
\subsection{Channel-wise Alignment}
In this section, we provide four types of channel-wise alignment operations.\par

\vspace{1mm}
\noindent\textbf{Dynamic Channel Projection:}
A straightforward way to align the dynamic width of the student with the teacher is to use a weight-sharing dynamic linear layer $W_i\in\mathcal{R}^{c^S_i\times c^T_i}$ \cite{YuYXYH19} to adjust the channel dimension of the sampled sub-network to be the same as the teacher model, where $g(f_i)$ is

\begin{equation}
    g (f_i)=\begin{cases}
    W_i^T\cdot f_i &, \text{if } f_i \text{ comes from student} \\
    f_i &, \text{if } f_i \text{ comes from teacher}
    \end{cases}
\end{equation}

\vspace{1mm}
\noindent\textbf{Spatial Attention:} 
Attention modules play a critical role in knowledge transfer.
In \cite{ZagoruykoK17}, it significantly improves the performance of the student by forcing it to mimic the attention maps of the teacher in image classification. 
Following \cite{ZagoruykoK17}, we conduct a non-local module \cite{0004GGH18} for each layer of extracted feature map of teacher model and student model, reducing the channel dimension to 4.

\vspace{1mm}
\noindent\textbf{Channel Maximize:} 
Another simple scheme is to compress the width of both student and teacher into one without any learnable parameters.
We provide two types of channel compression methods: maximize and average. 
The Channel Maximize operation selects a channel with maximum activation for each feature point along the spatial dimension ($x$, $y$), 
where $g (f_i)$ can be expressed as
\begin{equation}
    g (f_i)_{x,y}= \max\limits_{k=1,2,...,c_i}|f_{k,x,y}|.
\end{equation}

\vspace{1mm}
\noindent\textbf{Channel Average:}
Instead of focusing on the maximum activated channel, the average operation estimates the global average, where $g(f_i)$ can be formulated as: 
\begin{equation}
    g (f_i)_{x,y}= \frac{1}{c_i} \sum_{k=1}^{c_i}|f_{k,x,y}|.
\end{equation}

\section{Experimental Results}
In this section, we provide experimental results on the Sintel \cite{sintel} and KITTI \cite{kitti} benchmark to demonstrate the effectiveness of FlowNAS. 
Our method achieves state-of-the-art accuracy-efficiency trade-offs. 
In addition, we conduct ablation studies to discuss the effect of search space selection and FAD.

\subsection{Implementation Details}

\vspace{1mm}\noindent\textbf{Basic Setups:} We use RAFT \cite{TeedD20raft} as our baseline method and closely follow the training schedule of RAFT \cite{TeedD20raft} if not specified. 
Specifically, we first pre-train the network on FlyingChairs \cite{dosovitskiy2015flownet} for 100k iterations and then on FlyingThings \cite{flyingthings} for another 100k iterations. 
After pre-training, we finetune the network for Sintel evaluation on the combination of FlyingThings  \cite{flyingthings}, Sintel \cite{sintel}, KITTI-2015 \cite{menze2015object} and HD1K \cite{hd1k} for 100k iterations. 
Finally, we finetune on KITTI-2015 for an additional 50k iterations for KITTI evaluation. 
%
%
%
%

The main evaluation metric we use is average end-point-error (AEPE), the mean pixelwise flow error. In addition, KITTI  uses the F1-all (\%) metric, which refers to the percentage of optical flow vectors whose end-point error is larger than 3 pixels or over 5\% of ground truth.

\par 

\vspace{1mm}\noindent\textbf{FlowNAS Setups:} 
To search for the architecture with the best generalization ability and avoid overfitting, we split Sintel and KITTI training sets into training and validation pairs following FlowNet~\cite{flownet} and VCN~\cite{vcn} respectively. 
In the following, \textit{the referred training set and validation set of Sintel and KITTI are the corresponding set after splitting. 
The original training set officially provided is denoted as trainval set}. 
For a fair comparison with other methods, we report the final results of super-network trained on the trainval set after selecting the best architecture. 
During the evolutionary search, the parameter upper bound $P$ is set to 5.3M in this paper. 
To prevent the search from falling into local minima, we combine the FAD loss and the prediction error of the sub-network as regularization terms.
For each iteration, we select the top 50 architectures for crossover and mutation to produce another 50 architectures. The maximum number of iterations is set to 20.
The entire architecture search optimization takes about 1.4 GPU days for Sintel and 0.25 GPU days for KITTI on an RTX 8000 GPU.

\setlength\tabcolsep{.7em}
\begin{table}[!t]
\begin{center}
\caption{\textbf{Results on Sintel and KITTI datasets.} C+T denotes results after pre-training on FlyingChairs (C) and FlyingThings (T). C+T+S+K+H indicates the results trained on the combination of Sintel, KITTI, and HD1K. *Results evaluated with the “warm-start” strategy mentioned in RAFT~\cite{TeedD20raft}. FlowNAS-RAFT-S/K denotes the sub-network searched on Sintel or KITTI.}
\label{table:result}
\resizebox{\textwidth}{!}{
\begin{tabular}{clccccccc}
\toprule
\multirow{2}{*}{Training Data} & \multirow{2}{*}{Method} & \multicolumn{2}{c}{\underline{Sintel (trainval)}} &  \multicolumn{2}{c}{\underline{KITTI-15 (trainval)}} & \multicolumn{2}{c}{\underline{Sintel (test)}} & \multicolumn{1}{c}{\underline{KITTI-15 (test)}} \\
& & Clean & Final & F1-epe & F1-all & Clean & Final & F1-all \\
\midrule    
                       - & FlowFields~\cite{flowfields}     & -     & -  & - & -  & 3.75  & 5.81 & 15.31 \\
                       - & FlowFields++~\cite{flowfields++}     & - & - & - & -  & 2.94 & 5.49 & 14.82 \\
                       S & DCFlow~\cite{dcflow}     & -     & - & - & -   & 3.54  & 5.12 & 14.86 \\
                       S & MRFlow~\cite{mrflow}         & -  & -  & - & - & 2.53  & 5.38 & 12.19 \\ \midrule
\multirow{13}{*}{C + T} 
                       & HD3~\cite{hd3}            & 3.84  & 8.77 & 13.17 & 24.0 & - & - & - \\
                       & PWC-Net~\cite{pwcnet}        & 2.55  & 3.93 & 10.35 & 33.7 & - & - & - \\
                       & LiteFlowNet2~\cite{liteflownet2}   & 2.24  & 3.78  & 8.97 & 25.9 & - & - & - \\
                       & VCN~\cite{vcn}            & 2.21  & 3.68  & 8.36 & 25.1 & - & -     & - \\ 
                       & MaskFlowNet~\cite{maskflownet} & 2.25 & 3.61 & - & 23.1 & - & - & - \\ 
                       & FlowNet2~\cite{ilg2017flownet}       & 2.02  & 3.54 & 10.08 & 30.0 & 3.96  & 6.02 & - \\
                       & RAFT~\cite{TeedD20raft}        & 1.43 & 2.71 & 5.04 & 17.4 & - & - & - \\ 
                       & SCV \cite{JiangLL021} &  1.29 & 2.95 &  6.80 &  19.3 & - & - & - \\ 
                       & SeparableFlow~\cite{Zhang2021SepFlow}        & 1.30 & 2.59 & 4.60 & 15.9 & - & - & - \\ 
                       & GMA~\cite{jiang2021gma}       & 1.30 & 2.74 & 4.69 & 17.1 & - & - & - \\ 
                       & AGFlow~\cite{agflow}            & 1.31 & 2.69 & 4.82 & 17.0 & - & - & - \\
                      \cmidrule[\lightrulewidth](r{0.1em}){2-9}
                       & FlowNAS-RAFT-S        & 1.31 & 2.68 & - & - & - & - & - \\ 
                       & FlowNAS-RAFT-K        & - & - & 4.88 & 17.1 & - & - & - \\ 
\midrule

\multirow{16}{*}{C+T+S+K+H}
                     & LiteFlowNet2$^2$~\cite{liteflownet2} & (1.30) & (1.62) & (1.47) & (4.8) & 3.48  & 4.69 & 7.74 \\
                     & PWC-Net+~\cite{pwcnet+}   & (1.71)     & (2.34)  & (1.50) & (5.3)     
                     & 3.45  & 4.60 & 7.72 \\
                     & VCN~\cite{vcn}            & (1.66)     & (2.24) & (1.16) & (4.1) & 2.81  & 4.40 & 6.30 \\
                     & MaskFlowNet~\cite{maskflownet} & - & - & - & - & 2.52 & 4.17 & 6.10 \\

                     & RAFT~\cite{TeedD20raft} & (0.77) & (1.27) & - & - & 1.61* & 2.86*  & - \\
                     & SCV \cite{JiangLL021} & (0.86) & (1.75) & - & - & 1.77* & 3.88* & -\\
                     & SeparableFlow~\cite{Zhang2021SepFlow} & (0.69) & (1.10) & (0.69) & (1.60) & 1.50 & 2.67 & 4.64 \\
                     & GMA~\cite{jiang2021gma} & (0.62) & (1.06) & (0.56) & (1.2) & 1.39* & 2.47* & 4.93\\
                     & AGFlow~\cite{agflow} & (0.65) & (1.07) & (0.58) & (1.2) & 1.43* & 2.47* & 4.89 \\
                     & GMFlow~\cite{xu2022gmflow} & - & -&-&-& 1.74 & 2.90 & 9.32
                     \\
                     \cmidrule[\lightrulewidth](r{0.1em}){2-9}
                     & Vanilla FlowNAS-RAFT-S & (0.66) & (1.12) & - & - & 1.60* & 2.74* & - \\ 
                     & Vanilla FlowNAS-RAFT-K & - & - & (0.90) & (2.42) & - & - & 4.98 \\ 
                     & FlowNAS-RAFT-S & (0.77) & (1.25) & - & - & 1.65*
 & 3.16* & - \\ 
                     & FlowNAS-RAFT-K & - & - & (0.81) & (2.30) & - & - & 4.67
                     \\
                     & FlowNAS-GMFlow-S & (0.71) & (1.30) & - & - & 1.68 & 2.83 & -
                     \\
                     & FlowNAS-GMFlow-K & - & - & (1.521) & (5.11) & - & - & 8.77
                     \\

\bottomrule
\end{tabular}
}
\end{center}
\end{table}

\subsection{Main Results}
\tabref{table:result} compares our approach on Sintel and KITTI-2015 with prior works. We perform an evolutionary algorithm on Sintel and KITTI validation sets to find the two most suitable sub-network architectures. The searched architectures are named FlowNAS-RAFT-S and FlowNAS-RAFT-K, respectively. 

\vspace{1mm}\noindent\textbf{Vanilla FlowNAS:} Compared with RAFT \cite{TeedD20raft}, Vanilla FlowNAS-RAFT has already achieved better results on both Sintel and KITTI-2015. 
We train the searched architecture from scratch under the same protocol of \cite{TeedD20raft}, the Final AEPE of Sintel is reduced from 2.86 to 2.74, and the F1-all of KITTI is reduced from 5.10\% to 4.98\%.  
The results demonstrate the necessity of redesigning the encoder for flow estimation and the effectiveness of Vanilla FlowNAS. \par

\vspace{1mm}\noindent\textbf{FlowNAS:} FlowNAS equipped with FAD further improves the accuracy of Vanilla FlowNAS without retraining the sub-network. 
In \tabref{table:result}, on the training set of FlyingChairs (C) + FlyingThings (T), our approach achieves an AEPE of 1.31 on clean pass of Sintel, which is competitive to GMA \cite{jiang2021gma} and AGFlow \cite{agflow}  
and lower than RAFT \cite{TeedD20raft} by 8.4\% (from 1.43 to 1.31). 
On the final pass, it obtains a score of 2.68 AEPE, outperforming previous state-of-the-art methods SCV  \cite{JiangLL021} and RAFT by 9\% (from 2.95 to 2.68) and 0.1\% (from 2.71 to 2.68), respectively. 
FlowNAS-RAFT-K achieves an AEPE of 4.88 and F1-all score of 17.1\% on the KITTI trainval set, which significantly surpass SCV \cite{JiangLL021} by 28.1\% (from 6.80 to 4.89) and 11.4\% (from 19.3 to 17.1), respectively. The results demonstrate the excellent cross dataset generalization of our model.

For online evaluation, on KITTI benchmark, FlowNAS improves RAFT by 8.4\% (from $5.10\%$ to $4.67\%$) on F1-all, surpassing the state-of-the-art handcrafted GMA \cite{jiang2021gma}, SCV \cite{JiangLL021} and AGFlow \cite{agflow}. 
%
We achieve comparable KITTI F1-all error with SeparableFlow while reducing the number of parameters from 6.0M to 5.2M.
On the synthetic dataset Sintel, FlowNAS gets a higher AEPE than RAFT. Note that to prevent overfitting, other methods usually use validation error to find the best iteration checkpoint, while we omit this step to maintain a simple searching procedure and reduce the search time. We can still prove that we can search for a better architecture by Vanilla NAS.
Overall, the results demonstrate the strong representation ability of FlowNAS. 

Besides, to demonstrate that FlowNAS can be easily incorporated with existing networks to find a better encoder, we apply FlowNAS to another flow estimator with different decoder type, GMFlow. We name it FlowNAS-GMFlow. FlowNAS-GMFlow improves GMFlow from 1.74 to 1.68 on clean pass and 2.90 to 2.83 on final pass for Sintel. On KITTI, FlowNAS-GMFlow reduces the F1-all of GMFlow, from 9.32 to 8.77. The results show that FlowNAS is general and can be incorporated with different flow estimators to further boost their performance. More results can be found in Appendix.

\subsection{Ablation Study}
\label{sec:ablation}
In this section, we conduct ablation experiments to verify our design for FlowNAS. 
Due to the submission limits of Sintel and KITTI, we mainly focus on the experiment results on our validation split. The baseline method is RAFT \cite{TeedD20raft}.\par

\vspace{1mm}
\noindent\textbf{Search Space:} 
Since we are the first to introduce NAS into the flow task, discussing the search space selection is necessary. 
In \tabref{table:search space}, we evaluate four search spaces: Conv, SepConv, ShuffleConv and the combination of them. 
For convenience, the kernel size is the only variable of each search space.
We randomly sample 6 architectures for each search space and report their average training-from-scratch results. The sub-network is trained on FlyingChairs (C) + FlyingThings (T) for efficiency.  \par

We observe that SepConv works best, and ShuffleConv is the least effective for flow estimation.
Thus, we choose SepConv as the basic convolution operation for FlowNAS.
Furthermore, we find that the combination works worse than SepConv.
The reasons are as follows: first, the combination search space is $6^3$ times larger than that of SepConv, leading to insufficient training of each operation; second, the coupling of SepConv and other operations brings difficulties for the evolutionary algorithm to converge into a better local minimum. 

\setlength\tabcolsep{.7em}
\begin{table}[!t]
\begin{center}
\caption{\textbf{Ablation experiments of search space.} The SepConv search space performs the best for flow estimation task.}
\label{table:search space}
\resizebox{\textwidth}{!}{
\begin{tabular}{clcccc}
\toprule
\multirow{2}{*}{Training Data}  & \multirow{2}{*}{Search Space} & \multicolumn{2}{c}{\underline{Sintel (trainval)}} &  \multicolumn{2}{c}{\underline{KITTI-15 (trainval)}} \\
& & Clean & Final & F1-epe & F1-all \\
\midrule    
\multirow{4}{*}{C + T}
                       & Conv            & 1.44$\pm$0.07  & \textbf{2.71}$\pm$0.11 & 5.11$\pm$0.21 & 17.4$\pm$0.41 \\
                       & SepConv      & \textbf{1.40}$\pm$0.05  & 2.73$\pm$0.10 & \textbf{4.94}$\pm$0.18 & \textbf{16.9}$\pm$0.32 \\
                       & ShuffleConv       & 1.49$\pm$0.09  & 2.73$\pm$0.13 & 5.51$\pm$0.24 & 17.9$\pm$0.47 \\
                       & Combination        & 1.47$\pm$0.10  & 2.74$\pm$0.13 & 5.33$\pm$0.27 & 17.8$\pm$0.44 \\
\bottomrule
\end{tabular}
}
\end{center}
\end{table}

\subsubsection{Feature Alignment Distillation (FAD):} We conduct the following experiments to verify our design for FAD. 
\textit{For convenience, the results of two sub-network architectures on Sintel and KITTI are listed in the same row.}\par 

\vspace{1mm}
\noindent\textbf{Channel-wise Alignment}: 
We train FlowNAS super-network with the proposed four types of alignment operations under a full training schedule and report the performances of their best-searched architectures by directly inheriting weights from super-networks. 
As shown in \tabref{table:operation}, the operations with learnable parameters (Dynamic Channel Projection and Spatial Attention) are less effective than non-parametric operations. 
We speculate that this is because their number of parameters is so large that they interfere with the training of super-network. 
For example, the parameters of the Non-local module are 17\% of those of the super-network and may overfit the encoder feature, which reduces the performance of the super-network.

\setlength\tabcolsep{.7em}
\begin{table}[!t]
\begin{center}
\caption{\textbf{Ablation experiments for Channel-wise Alignment operation.} The Channel Maximize operation achieves the best result.}
\label{table:operation}
\resizebox{\textwidth}{!}{
\begin{tabular}{clcccc}
\toprule
\multirow{2}{*}{Training Data} & \multirow{2}{*}{feature alignment Operation} & \multicolumn{2}{c}{\underline{Sintel (val)}} &  \multicolumn{2}{c}{\underline{KITTI-15 (val)}} \\
& & Clean & Final & F1-epe & F1-all \\
\midrule    
\multirow{4}{*}{C+T+S (train)+K (train)+H} 
                       & Dynamic Channel Projection     & 1.88 & 5.06 & 1.75 & 4.80 \\
                       & Spatial Attention     & 22.3 & 30.1 & 26.8 & 83.0 \\
                       & Channel Maximize     & \textbf{1.16} & \textbf{3.00} & \textbf{1.31} & \textbf{3.63} \\
                       & Channel Average      & 1.20 & 3.22 & 1.41 & 3.89 \\
\bottomrule
\end{tabular}
}
\end{center}
\end{table}

\begin{figure}[!t]
\caption{KITTI-15 validation results of several random sub-network architectures, whose weights come from Vanilla FlowNAS, From-scratch, and FlowNAS.}
	\begin{center}
	    \subfigure[Vanilla FlowNAS vs. FlowNAS]{
	\includegraphics[width=0.42\linewidth]{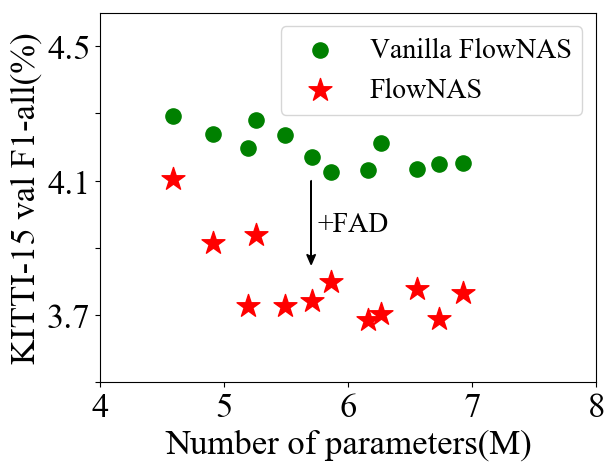}
	}
	\subfigure[FlowNAS vs. From-scratch]{
	\includegraphics[width=0.42\linewidth]{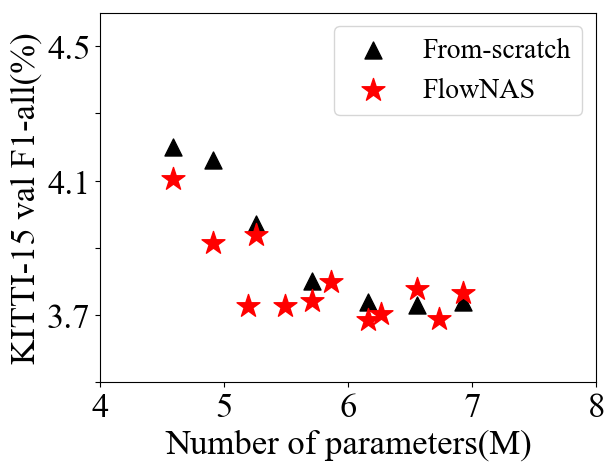}
	}
	\end{center}
	\vspace{-10pt}
	\label{fig:distill}
\end{figure}

\vspace{1mm}
\noindent\textbf{Effectiveness of FAD:} 
\figref{fig:distill}.(a) compares the performance of super-network trained with or without FAD.
We randomly sample 12 sub-networks and report their performance with inherited weights from Vanilla FlowNAS super-network and FlowNAS super-network, respectively. For all sampled architectures, FlowNAS reduces their F1-all error by 0.18$\sim$0.51, verifying the effectiveness of FAD for better optimizing super-network.
\figref{fig:distill}.(b) compares the performance of sub-networks derived from super-network or trained from scratch. We randomly sample 6 sub-networks and train them from scratch as their stand-alone performance. For all sampled architectures, super-network weights achieve similar or even lower F1-all error than their stand-alone performance, showing the strong representation ability of FlowNAS.

\tabref{table:distillation} compares the performance of the two best architectures searched by Vanilla FlowNAS and FlowNAS, namely Vanilla FlowNAS-RAFT and FlowNAS-RAFT.  Line 1 and 3 show their performance when inheriting weights from their corresponding super-networks. FlowNAS-RAFT achieves a lower F1-all error of 0.42 than Vanilla FlowNAS-RAFT, further validating the effectiveness of FAD for super-network training. 
Finally, line 4 and 5 compare the performance of FlowNAS-RAFT when trained with or without FAD. We surprisingly find that FAD can also improve the training of sub-network by reducing 0.06 F1-all error, further validating the effectiveness of FAD for flow estimator training.

\setlength\tabcolsep{.7em}
\begin{table}[t]
\begin{center}
\caption{\textbf{Ablation experiments for distillation.} The entry ``from-scratch*'' denotes training the sub-network with Feature Alignment Distillation.}
\label{table:distillation}
\resizebox{\textwidth}{!}{
\begin{tabular}{clccc}
\toprule
\multirow{2}{*}{Training Data} & \multirow{2}{*}{Method} & \multirow{2}{*}{Weight}  &  \multicolumn{2}{c}{\underline{KITTI-15 (val)}} \\
& &  &  F1-epe &  F1-all \\
\midrule    
\multirow{5}{*}{C+T+S (train)+K (train)+H} 
                       & Vanilla FlowNAS-RAFT      & inherit   & 1.52 & 4.05 \\
                       & Vanilla FlowNAS-RAFT      & from-scratch   & 1.36 & 3.69 \\
                       & FlowNAS-RAFT      & inherit   & \textbf{1.31} & \textbf{3.63} \\
                       & FlowNAS-RAFT      & from-scratch   & 1.38 & 3.84 \\
                       & FlowNAS-RAFT      & from-scratch*   & 1.36 & 3.78 \\
\bottomrule
\end{tabular}
}
\end{center}
\end{table}



We also refine the initial features of super-network following the existing dense image prediction method \cite{ZhaoSQWJ17}. 
We simplify the decoder by directly estimating the optical flow based on the similarities between two pyramid feature maps from super-network, namely rough decoder. 
The supervision loss of rough decoder is denoted as $\mathcal{L}_{rough}$. 
As shown in \tabref{table:loss}, unlike the additional loss of other tasks, e.g., semantic segmentation, we find that our super-network performs worse or even collapses when using rough decoder as supervision. 
The rough decoder has no learnable parameter and cannot give an accurate optical flow estimation, in contrast to the standard flow estimator (decoder) that can progressively obtain refined flow estimation based on two pyramids of features and cost volumes.


\setlength\tabcolsep{.7em}
\begin{table}[!t]
\begin{center}
\caption{\textbf{Ablation experiments for additional supervision.} Rough decoder does not help refine super-network.} 
\label{table:loss}
\resizebox{\textwidth}{!}{
\begin{tabular}{clcccc}
\toprule
\multirow{2}{*}{Training Data} & \multirow{2}{*}{Supervision Loss} & \multicolumn{2}{c}{\underline{Sintel (val)}} &  \multicolumn{2}{c}{\underline{KITTI-15 (val)}} \\
& & Clean & Final & F1-epe & F1-all \\
\midrule    
\multirow{4}{*}{C+T+S (train)+K (train)+H} 
                       & $\mathcal{L}_{flow}$     & 1.24 & 3.30 & 1.52 & 4.05 \\
                       & $\mathcal{L}_{flow}+\mathcal{L}_{rough}+\mathcal{L}_{D}$     & 1.56 & 3.91 & 1.50 & 3.94 \\
                       & $\mathcal{L}_{rough}+\mathcal{L}_{D}$      & 30.8 & 39.8 & 32.3 & 98.7 \\
                       & $\mathcal{L}_{flow}+\mathcal{L}_{D}$      & \textbf{1.16} & \textbf{3.00} & \textbf{1.31} & \textbf{3.63} \\
\bottomrule
\end{tabular}
}
\end{center}
\end{table}

\setlength\tabcolsep{.7em}
\begin{table}[!t]
\begin{center}
\caption{\textbf{Overall comparison with cutting edge handcrafted flow estimators.} FlowNAS enables the most advanced accuracy-efficiency trade-off.}
\label{table:parameters}
\begin{tabular}{cccccc}
\toprule
\multirow{2}{*}{Method} & \multirow{2}{*}{Params} & \multirow{2}{*}{GFLOPs} & \multirow{2}{*}{Time} & \underline{KITTI (test)} \\
& & & & Fl-all (\%) \\
\midrule    
PWCNet+~\cite{pwcnet+} & 9.4M & \textbf{90.8} & \textbf{0.02} & 7.72 \\
VCN\cite{vcn} & 6.2M & 96.5 & 0.11 & 6.30 \\
RAFT~\cite{TeedD20raft} & 5.3M & 388 & 0.17  & 5.10 \\
GMA~\cite{jiang2021gma} & 5.9M & 435 & 0.20 & 4.93 \\
SeparableFlow~\cite{Zhang2021SepFlow} & 6.0M & 495 & 0.23 & \textbf{4.63} \\
\midrule
FlowNAS-RAFT-K & \textbf{5.2M}  & 368 & 0.19 & 4.67 \\
\bottomrule
\end{tabular}
\end{center}
\end{table}

\subsection{Parameter, Timing and Accuracy}
\label{params}
FlowNAS improves the performance of the baseline and reduces the number of parameters, inference latency, and GFLOPs. 
We measure the inference time of existing flow networks on the same machine with 1 RTX 8000 GPU. 
We set the input size as 384$\times$1280 for GFLOPs calculation and inference time. 
The iterations of RAFT \cite{TeedD20raft}, GMA \cite{jiang2021gma}, SperarableFlow \cite{Zhang2021SepFlow}, and FlowNAS is set to 24.
%
%
The results are shown in \tabref{table:parameters} and \figref{fig:time}.
FlowNAS achieves the best accuracy-efficiency trade-offs.
We achieve a comparable result with SperarableFlow, while reducing 13\% parameters and 26\% FLOPs.
Moreover, we joint search the encoder architecture and iteration number of the decoder to obtain a better Pareto front of accuracy and GFLOPs. More results can be found in Appendix.


\section{Conclusions}
In this work, we address the problem of automatically designing the encoder of flow estimators. 
We propose FlowNAS to find the optimal encoder structure specifically for flow estimators.
We study different search spaces and construct the super-network on the best search space for neural architecture search. 
To improve the accuracy of super-network and remove the retraining stage of sub-network, we propose Feature Alignment Distillation, which guides the training of all the sub-network of super-network. 
The proposed FlowNAS can be easily incorporated with any existing networks to find a better encoder.
Experimental results show that FlowNAS achieves the state-of-the-art accuracy-efficiency trade-offs on the challenging Sintel and KITTI benchmarks. We plan to extend FlowNAS to search for the optimal encoder and decoder structure for flow estimation in the future.



\clearpage

%
%
\bibliographystyle{splncs04}
\bibliography{egbib}

\clearpage

\def\x{{\mathbf x}}
\def\L{{\cal L}}

\renewcommand\thefigure{S\arabic{figure}}
\renewcommand\thetable{S\arabic{table}}
\setcounter{section}{0}
\setcounter{figure}{0}
\setcounter{table}{0}

\section*{Appendix}

\section{More Results of FlowNAS}
We test Vanilla FlowNAS and FlowNAS on Sintel test-dev without warm-start trick. The results are shown in \tabref{table:nowarmresult}. Vanilla FlowNAS surpasses RAFT~\cite{TeedD20raft} by 0.26 (from 1.94 to 1.68) on Clean and 0.08 (from 3.18 to 3.10) on Final. \par

\setlength\tabcolsep{.7em}
\begin{table}[!ht]
\begin{center}
\caption{\textbf{Results on Sintel test-dev.} C+T+S+K+H indicates the results are trained on the combination of Sintel, KITTI, and HD1K. The results are reported without using the warm-start trick in RAFT~\cite{TeedD20raft}.}
\label{table:nowarmresult}
\resizebox{0.8\textwidth}{!}{
\begin{tabular}{clcccc}
\toprule
\multirow{2}{*}{Training Data} & \multirow{2}{*}{Method} & \multicolumn{2}{c}{\underline{Sintel (trainval)}} &  \multicolumn{2}{c}{\underline{Sintel (test)}} \\
& & Clean & Final & Clean & Final \\
\midrule    
\multirow{3}{*}{C+T+S+K+H}
                     
                      & RAFT~\cite{TeedD20raft} & (0.76) & (1.22) &  1.94 & 3.18   \\
                     & Vanilla FlowNAS-RAFT-S & (0.66) & (1.12) & 1.68 & 3.10\\ 
                     & FlowNAS-RAFT-S & (0.77) & (1.25) & 1.93 & 3.38 \\ 

\bottomrule
\end{tabular}
}
\end{center}
\end{table}

\begin{figure}[!ht]
	\centering
	\caption{\textbf{Trade-offs on accuracy and GFLOPs}. FlowNAS obtain a better Pareto front.
	}
	\includegraphics[width=1.0\linewidth]{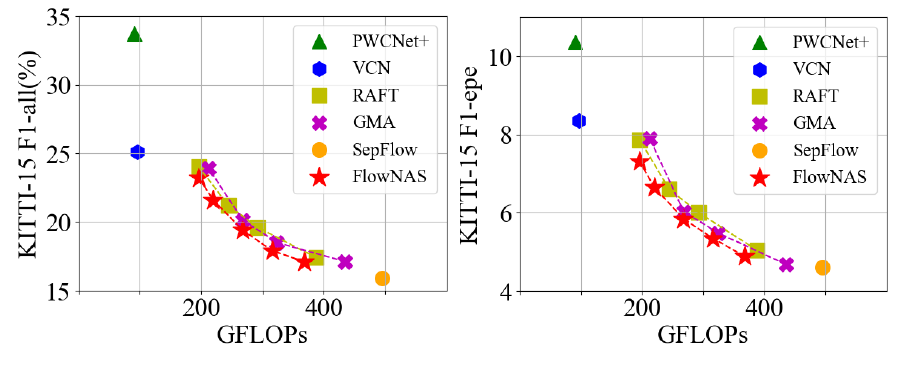}
	\label{fig:flops}
\end{figure}

In addition, to verify that FlowNAS achieves better accuracy-GFLOPs trade-offs than handcrafted flow estimator, we jointly search the architecture and iteration number of FlowNAS under different GFLOPs, and draw the Pareto front, as shown in \figref{fig:flops}. We adjust the iteration number of RAFT and GMA~\cite{jiang2021gma} to obtain their Pareto fronts. For convenience, we report the KITTI results when models are trained on C+T training set. We can find that FlowNAS achieves a better Pareto front of accuracy and GFLOPs than RAFT~\cite{TeedD20raft} and GMA~\cite{jiang2021gma}.  \par

To demonstrate that FlowNAS can be easily incorporated with existing networks to find a better encoder,
we apply FlowNAS to a better flow estimator, GMA~\cite{jiang2021gma}. We name it FlowNAS-GMA, as shown in \tabref{table:gma}. For Sintel, FlowNAS-GMA reduces AEPE by 0.04 (from 1.30 to 1.26) on Clean and 0.18 (from 2.74 to 2.56) on Sintel. On KITTI-2015, the error rate is reduced from 1.71 to 16.9 by FlowNAS-GMA. The results show that, even with a stronger flow estimator, FlowNAS can further boost the performance. \par

\setlength\tabcolsep{1.7em}
\begin{table}[!ht]
\begin{center}
\caption{\textbf{Results of FlowNAS-GMA.} C+T denotes results after pre-training on FlyingChairs (C) and FlyingThings (T). FlowNAS-GMA achieves a lower error rate than GMA~\cite{jiang2021gma} while reducing the GFLOPs number.} 
\label{table:gma}
\resizebox{\textwidth}{!}{
\begin{tabular}{clcccc}
\toprule
\multirow{2}{*}{Training Data} & \multirow{2}{*}{Method} & \multicolumn{2}{c}{\underline{Sintel (trainval)}} &  \multicolumn{2}{c}{\underline{KITTI-15 (trainval)}} \\
& & Clean & Final & F1-epe & F1-all \\
\midrule    
\multirow{3}{*}{C + T}
                     
                      & GMA~\cite{TeedD20raft} &  1.30 & 2.74 & 4.69 & 17.1    \\
                     & FlowNAS-GMA-S & \textbf{1.26} & \textbf{2.56} & - & - \\ 
                     & FlowNAS-GMA-K & - & - & \textbf{4.66} &\textbf{16.9} \\ 

\bottomrule
\end{tabular}
}
\end{center}
\end{table}

Besides working on the flow estimators with the recurrence module like RAFT and GMA, we also apply FlowNAS to a more efficient model, PWC-Net. For convenience, we directly apply searched encoder
by RAFT on PWC-Net\cite{pwcnet}. The results are shown in \tabref{table:pwc}. Our searched encoder surpasses original PWC-Net by 28.2\% (1.83 v.s. 2.55) and 12.4\% (3.44 v.s. 3.93) on Sintel clean and final, showing the strong generalization ability of FlowNAS. 
\par

\setlength\tabcolsep{1em}
\begin{table}[!ht]
\begin{center}
\caption{\textbf{Results of PWC-Net with our searched encoder.} C+T denotes results after pre-training on FlyingChairs (C) and FlyingThings (T).} 
\label{table:pwc}
\resizebox{0.7\textwidth}{!}{
\begin{tabular}{cllcc}
\toprule
\multirow{2}{*}{Training Data} & \multirow{2}{*}{Encoder} & \multirow{2}{*}{Decoder} & \multicolumn{2}{c}{\underline{Sintel (trainval)}} \\
& & & Clean & Final \\
\midrule    
                     
C + T& PWC-Net~\cite{pwcnet} & \multirow{3}{*}{PWC-Net} &  2.55 & 3.93    \\
AutoFlow & PWC-Net && 2.17 & \textbf{2.91} \\ 
C + T & FlowNAS-RAFT-S && \textbf{1.83} & 3.44 \\ 

\bottomrule
\end{tabular}
}
\end{center}
\end{table}

\section{Details of Search Space}
\begin{figure}[!ht]
	\centering
	\caption{\textbf{Details of SepConv.} Each convolution is followed by a normalization layer and an activation layer.
	}
	\includegraphics[width=0.4\linewidth]{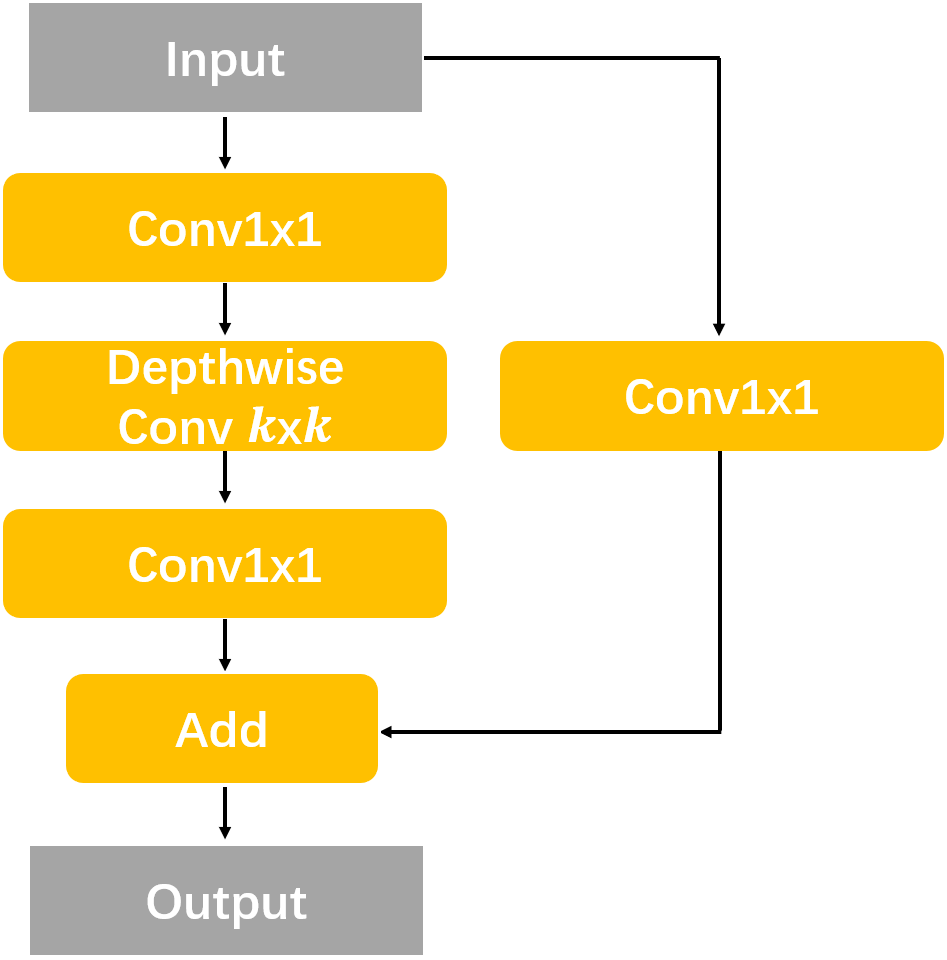}
	\label{fig:block}
\end{figure}

Our search space is defined in \tabref{table:search space detail}. The search space design is adapted from AttentiveNAS~\cite{Wang0GC21}. The stride of each block follows  the original encoder of RAFT. The detail of SepConv is shown in \figref{fig:block}. 

\setlength\tabcolsep{.7em}
\begin{table}[!t]
\begin{center}
\caption{\textbf{Search Space of FlowNAS.} Width refers to the number of channels. Depth is the number of repeated blocks. The expansion ratio denotes the channel expansion ratio of the first 1x1 convolution of SepConv.}
\label{table:search space detail}
\resizebox{\textwidth}{!}{
\begin{tabular}{cccccc}
\toprule
Block & Width & Depth & Kernel Size & Expansion Ratio & Stride\\
\midrule    
First~Conv2d    & \{64\} & \{1\} & \{7\} & - & 2\\
SepConv-1 & \{56,64\} & \{1,2\} & \{3,5\} & \{1\} & 1\\
SepConv-2 & \{64,72\} & \{1,2,3\} & \{3,5\} & \{1,2,4\} & 1\\
SepConv-3 & \{88,96\} & \{1,2,3\} & \{3,5\} & \{4,5,6\} & 2\\
SepConv-4 & \{96,104,112\} & \{1,2,3\} & \{3,5\} & \{4,5,6\} & 1\\
SepConv-5 & \{112,120,128\} & \{2,3,4\} & \{3,5\} & \{6\} & 2\\
SepConv-6 & \{128,136\} & \{1,2\} & \{3,5\} & \{6\} & 1\\
Last~Conv2d    & \{128\} & \{1\} & \{1\} & - & 1\\

\bottomrule
\end{tabular}
}
\end{center}
\end{table}

\section{Architecture Configuration of Sub-network}
We list the searched architecture configuration of sub-network on Sintel and KITTI, as shown in \tabref{table:subnet sintel} and \tabref{table:subnet kitti}.

\setlength\tabcolsep{.7em}
\begin{table}[!t]
\begin{center}
\caption{\textbf{Architecture configuration of sub-network searched on Sintel.}}
\label{table:subnet sintel}
\resizebox{\textwidth}{!}{
\begin{tabular}{cccccc}
\toprule
Block & Width & Depth & Kernel Size & Expansion Ratio & Stride\\
\midrule    
First~Conv2d    & 64 & 1 & 7 & - & 2\\
SepConv-1 & 64 & 2 & 3 & 1 & 1\\
SepConv-2 & 72 & 2 & 3 & 2 & 1\\
SepConv-3 & 96 & 2 & 5 & 5 & 2\\
SepConv-4 & 104 & 1 & 3 & 5 & 1\\
SepConv-5 & 120 & 2 & 5 & 6 & 2\\
SepConv-6 & 136 & 1 & 5 & 6 & 1\\
Last~Conv2d    & 128 & 1 & 1 & - & 1\\

\bottomrule
\end{tabular}
}
\end{center}
\end{table}

\setlength\tabcolsep{.7em}
\begin{table}[!t]
\begin{center}
\caption{\textbf{Architecture configuration of sub-network searched on KITTI.}}
\label{table:subnet kitti}
\resizebox{\textwidth}{!}{
\begin{tabular}{cccccc}
\toprule
Block & Width & Depth & Kernel Size & Expansion Ratio & Stride\\
\midrule    
First~Conv2d    & 64 & 1 & 7 & - & 2\\
SepConv-1 & 64 & 2 & 3 & 1 & 1\\
SepConv-2 & 72 & 1 & 5 & 4 & 1\\
SepConv-3 & 88 & 1 & 3 & 6 & 2\\
SepConv-4 & 104 & 2 & 5 & 5 & 1\\
SepConv-5 & 120 & 2 & 5 & 6 & 2\\
SepConv-6 & 136 & 1 & 5 & 6 & 1\\
Last~Conv2d    & 128 & 1 & 1 & - & 1\\

\bottomrule
\end{tabular}
}
\end{center}
\end{table}

\section{Visualization}
\begin{figure}[!ht]
	\centering
	\caption{Visualization results on the KITTI 2015 test-dev. The left column is the input image. The middle column is the estimated flow and error maps of FlowNAS. The right column is the estimated flow and error maps of RAFT baseline. From blue to red, the error of the estimated flow increases in the error map.
	}
    \begin{minipage}{1.0\linewidth}
	\includegraphics[width=1.0\linewidth]{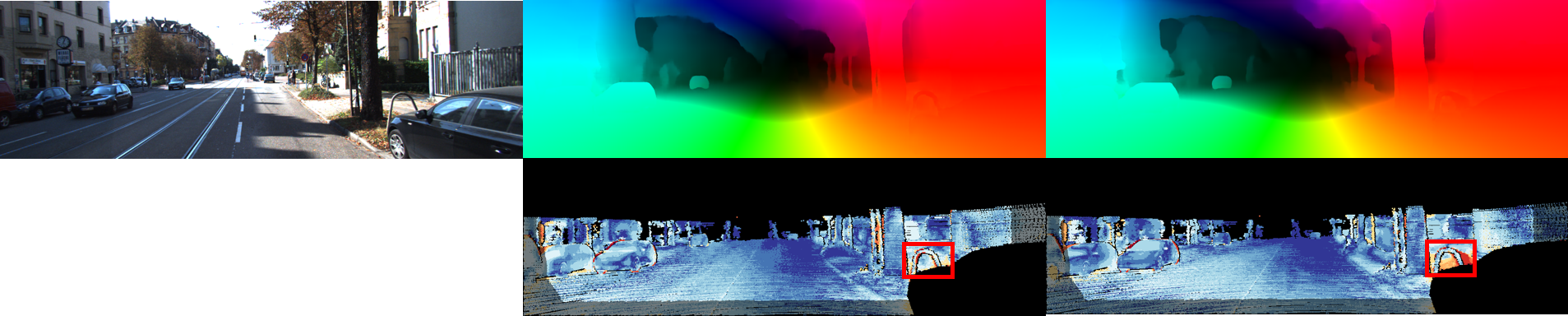} 
	\includegraphics[width=1.0\linewidth]{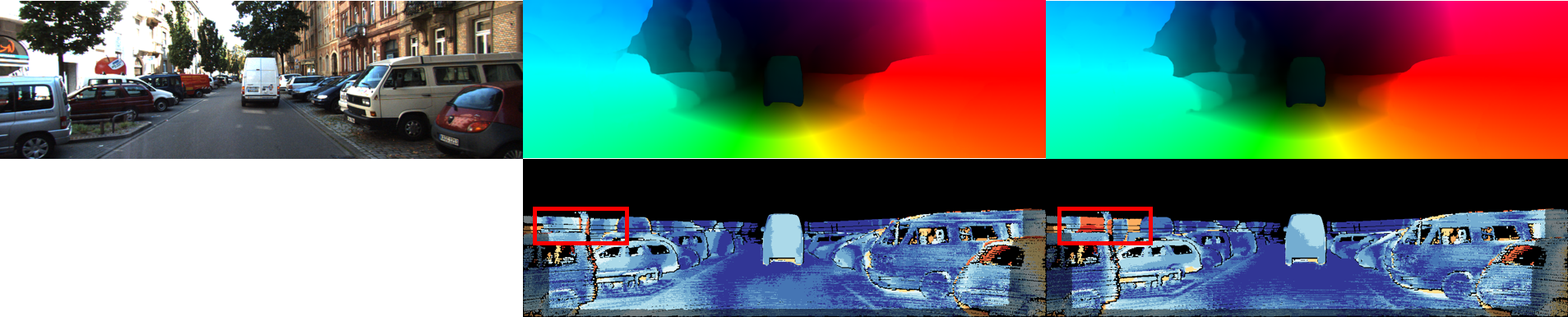} 
	\includegraphics[width=1.0\linewidth]{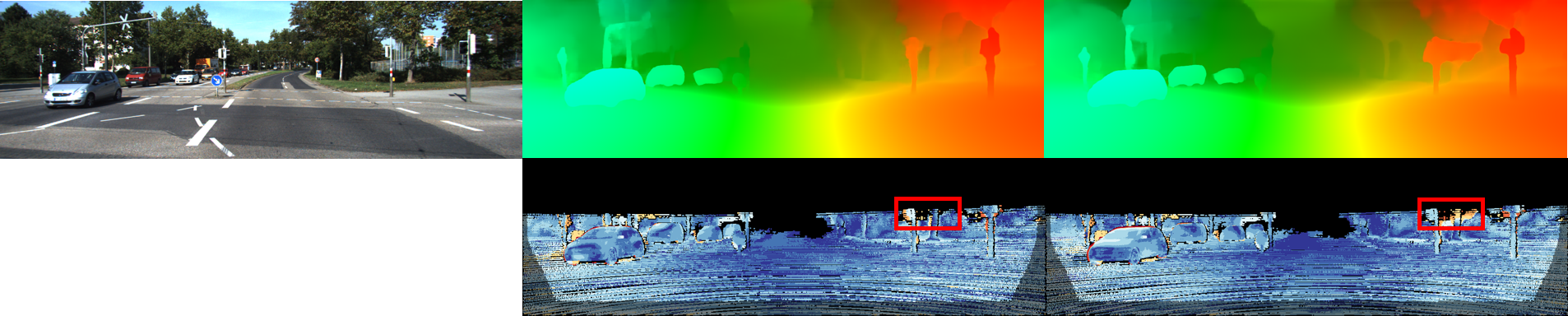} 
	\includegraphics[width=1.0\linewidth]{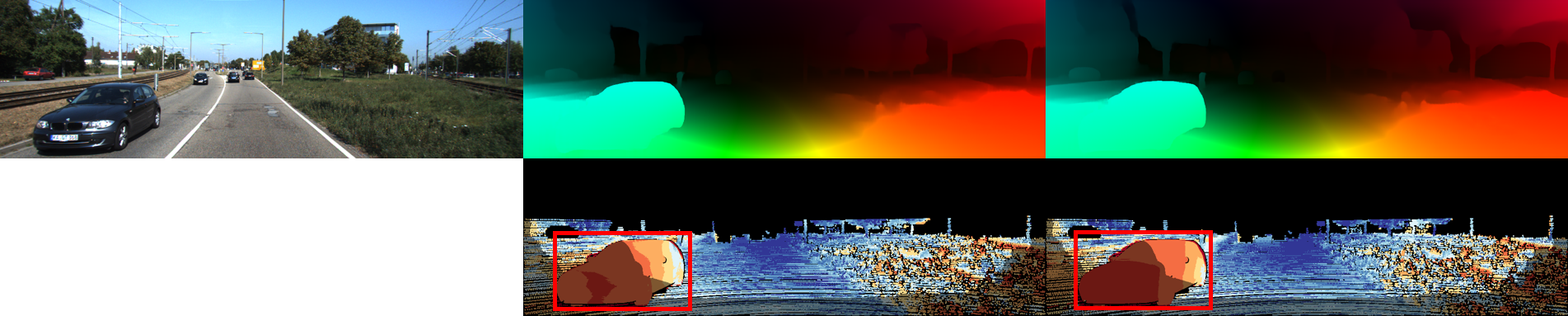}
    \end{minipage}
	\label{fig:kitti visualization}
\end{figure}

We compare the flow results of FlowNAS and RAFT on the KITTI test-dev in \figref{fig:kitti visualization}. From left to right, the figures are input images, flow visualization, and error maps of FlowNAS and those of RAFT. We can observe that, with a better encoder, FlowNAS improves the flow details of the background. Besides, for a fast-moving object (the car in the last row of \figref{fig:kitti visualization}), FlowNAS can capture more accurate movement than RAFT.

\end{document}